\documentclass[letterpaper, 10 pt, conference]{ieeeconf}  % Comment this line out if you need a4paper

\IEEEoverridecommandlockouts                              % This command is only needed if 
\usepackage{graphicx}
\usepackage{amsmath} % assumes amsmath package installed
\usepackage{amssymb}  % assumes amsmath package installed
\usepackage{xcolor}
\usepackage[hidelinks]{hyperref}
\usepackage[inkscapearea=page]{svg}
\usepackage{threeparttable}
\usepackage{listings}
\usepackage{stfloats}

\usepackage{fancyhdr}

\usepackage{titlesec}
\titlespacing*{\subsubsection}{0pt}{0.1\baselineskip}{0.2\baselineskip}
\titlespacing*{\subsection}{0pt}{0.2\baselineskip}{0.2\baselineskip}
\titlespacing*{\section}{0pt}{0.4\baselineskip}{0.2\baselineskip}

\newcommand\library[0]{Graphite}
\definecolor{codegreen}{rgb}{0,0.6,0}
\definecolor{codegray}{rgb}{0.5,0.5,0.5}
\definecolor{codepurple}{rgb}{0.58,0,0.82}
\definecolor{backcolour}{rgb}{0.95,0.95,0.92}
\definecolor{codeblue}{rgb}{0.306,0.788,0.690}

\lstdefinestyle{cuda}{
    commentstyle=\color{codegreen},
    keywordstyle=\color{blue},
    numberstyle=\tiny\color{codegray},
    stringstyle=\color{codepurple},
    basicstyle=\ttfamily\tiny,
    breakatwhitespace=false,         
    breaklines=true,                 
    captionpos=b,                    
    keepspaces=true,                 
    numbers=left,                    
    numbersep=5pt,                  
    showspaces=false,                
    showstringspaces=false,
    showtabs=false,                  
    tabsize=2,
    frame=lines,
    columns=fullflexible,
    xleftmargin=2.0em
}

\title{\LARGE \bf
\library{}: A GPU-Accelerated Mixed-Precision Graph Optimization Framework}
\author{Shishir Gopinath$^{1}$, Karthik Dantu$^{2}$, and Steven Y. Ko$^{1}$
\thanks{$^{1}$Simon Fraser University}
\thanks{$^{2}$University at Buffalo}
}

\newcommand\copyrighttext{%
  \scriptsize © 2026 IEEE.  Personal use of this material is permitted.  Permission from IEEE must be obtained for all other uses, in any current or future media, including reprinting/republishing this material for advertising or promotional purposes, creating new collective works, for resale or redistribution to servers or lists, or reuse of any copyrighted component of this work in other works.
  }

\begin{document}

\maketitle
\thispagestyle{empty}
\pagestyle{empty}
% \copyrightnotice
\thispagestyle{fancy}

%%%%%%%%%%%%%%%%%%%%%%%%%%%%%%%%%%%%%%%%%%%%%%%%%%%%%%%%%%%%%%%%%%%%%%%%%%%%%%%%
\begin{abstract}

We present \library{}, a GPU-accelerated nonlinear least squares graph optimization framework.
It provides a CUDA C++ interface to enable the sharing of code between a real-time application, such as a SLAM system, and its optimization tasks. The framework supports techniques to reduce memory usage, including in-place optimization, support for multiple floating point types and mixed-precision modes, and dynamically computed Jacobians. 
We evaluate \library{} on well-known bundle adjustment problems and find that it achieves similar performance to MegBA, a solver specialized for bundle adjustment, while maintaining generality and using less memory. We also apply \library{} to global visual-inertial bundle adjustment on maps generated from stereo-inertial SLAM datasets, and observe speed-ups of up to $59\times$ compared to a CPU baseline.
Our results indicate that our framework enables faster large-scale optimization on both desktop and resource-constrained devices. 

\end{abstract}

%%%%%%%%%%%%%%%%%%%%%%%%%%%%%%%%%%%%%%%%%%%%%%%%%%%%%%%%%%%%%%%%%%%%%%%%%%%%%%%%
\section{Introduction}

Nonlinear optimization is a key component of many estimation problems in computer vision, graphics, and robotics. It is especially prevalent in keyframe-based simultaneous localization and mapping (SLAM) systems, which construct a representation of the environment while also determining a device's position and rotation. In particular, ORB-SLAM3~\cite{ORBSLAM3}, a visual-inertial SLAM system, widely employs nonlinear optimization over hypergraphs for a broad array of frontend and backend tasks across its tracking, local mapping, and loop closing threads.

For real-time SLAM, it is necessary to perform these optimizations in a timely manner to remain up to date with changes in the device's state and surroundings. At the same time, the computational time and amount of memory needed to carry out each optimization may increase as the size of the map grows. This may be especially challenging for resource-constrained platforms when on-device SLAM is just one component of a larger system with several concurrent tasks.

Popular nonlinear optimizers~\cite{gtsam, g2o, Agarwal_Ceres_Solver_2022} take advantage of multi-core CPU architectures, exploiting the inherent parallelism of sparse linear algebraic operations. However, some platforms used for SLAM~\cite{Jeon2021} are equipped with on-board accelerators such as GPUs, which can be used to perform a variety of learning-based perception tasks efficiently, such as feature extraction, depth-estimation, object detection, and place recognition~\cite{Mokssit2023}. GPUs excel at these compute-heavy workloads due to their massively parallel architecture, which also makes them well-suited for nonlinear optimization. 

To date, there have been several efforts to speed up existing optimization libraries by identifying and offloading expensive steps to GPUs. More recently, Ceres Solver~\cite{Agarwal_Ceres_Solver_2022} implemented GPU acceleration for the linear solver step of trust region optimization algorithms. Other work based on g$^2$o~\cite{g2o} focuses on efficiently reducing the size of the linear system to be solved~\cite{GDK2023} and accelerating numerical differentiation for computing Jacobian matrices~\cite{Kumar2024}.
Yet, while it is possible to improve the performance of existing libraries by offloading individual steps to the GPU, doing so introduces additional overhead from allocating GPU memory and transferring data between steps.

In contrast, new GPU-based frameworks have emerged which accelerate various steps of the optimization.
Some use Python~\cite{OpenOF} or a domain-specific language (DSL)~\cite{Devito2017, Mara2021} to allow users to write simple descriptions of nonlinear optimization problems that are used to automatically generate efficient solvers using complex code transformations.
Others leverage machine learning frameworks to support batched optimization~\cite{Huang2021} and differentiable optimization~\cite{Yi2021, pineda_theseus_2023, wang2023pypose} for training models for tasks such as feature extraction and feature matching.
In addition, many solvers are specialized for specific classes of problems such as bundle adjustment~\cite{2021megba}, achieving large speedups. 

However, existing GPU-accelerated solvers have several limitations which make them challenging to adapt for applications such as SLAM.
They often model optimizable variables as simple numeric data types (e.g. a float, double, vector of doubles, etc.) or only provide data types for a specific optimization problem (e.g. a class representing an SE(3) transformation). Meanwhile, SLAM systems may represent variables as complex classes which consist of other classes. 
For example, an optimizable pose class may consist of multiple SE(3) transformations and camera references~\cite{ORBSLAM3}. 
Existing solvers cannot represent these classes with a vector of numbers or predetermined data types because they cannot model their complex behaviours and data dependencies.
Additionally, researchers develop real-time SLAM systems in a specific language of their choice, e.g., C++, making it difficult to use DSLs or Python-based optimizers, since they require optimizable data types and functionality to be faithfully reimplemented for another language.
A third challenge is that GPU memory is often limited and shared between multiple tasks such as image processing and model inference.
Moreover, existing libraries may demand additional GPU memory for operations on sparse matrices and linear solver workspace allocations~\cite{cuda}.
Even worse, embedded platforms often have no dedicated GPU memory, so the CPU and GPU must compete for the same memory.

In this paper, we present \library{}, a GPU-accelerated nonlinear least squares graph optimization framework, which implements several techniques for balancing runtime performance and GPU memory usage, to enable large-scale optimization for desktop and embedded scenarios.
It allows for mixed-precision solving using 64-bit, 32-bit, and 16-bit floating-point precisions, enabling faster and more memory-efficient optimization of graphs.
To further reduce memory usage, the library supports dynamically computed Jacobians for matrix-free methods, as well as an automatic differentiation method with equivalent memory overhead to analytic differentiation. In addition, the iterative linear solver is aware of the structure of the graph, bypassing explicit sparse matrix formats and their associated memory costs.
To support real-time SLAM and odometry, the framework provides a CUDA C++ interface, and uses a batching model which supports in-place optimization, allowing users to optimize GPU-accessible data without first transforming it into a solver-specific format.
This avoids memory usage and runtime overhead from unnecessary copying, as well as explicit data transfers on platforms which share CPU and GPU memory.

To demonstrate the effectiveness and flexibility of our framework, we evaluate it on bundle adjustment problems, and find it performs comparably to MegBA~\cite{2021megba} while using up to 78\% less GPU memory.
We also reimplement visual-inertial bundle adjustment on the GPU inside ORB-SLAM3, which can take hundreds of seconds on the CPU for sufficiently large problems, and demonstrate a speed up of up to $59 \times$ on maps generated by processing stereo-inertial SLAM datasets~\cite{schubert2018vidataset}, while using under 1 GiB of GPU memory on a desktop machine. The code is available at 
\url{https://github.com/sfu-rsl/graphite}.

To summarize our contributions:

\begin{itemize}
    \item We design and implement a general mixed-precision optimization framework for on-device estimation problems which performs all major optimization steps on a GPU.
    \item We apply our framework to global visual-inertial bundle adjustment inside a SLAM system, which consists of 7 types of constraints and 5 types of variables.
    \item We evaluate our framework across different precisions and differentiation modes, using well-known bundle adjustment and SLAM datasets.
\end{itemize}

\section{Background} \label{sec:background}

In this section, we provide a brief discussion on nonlinear optimization over constraints, the Levenberg-Marquardt algorithm implemented by our framework, visual-inertial optimization, and relevant GPU programming concepts.

\noindent\textbf{Optimizing Over Constraints:}
Some estimation problems can be modelled as an optimization over constraints. Each constraint represents a relationship between estimated parameters, an observation, and the error. The goal is then to find the parameters that minimize the total error across all constraints ($\mathcal{C}$). This may be formulated as 

\begin{equation} \label{eq:total-error}
    \footnotesize
    \text{arg}\min\limits_{x} \frac{1}{2}\sum_{\langle i, j \rangle \in \mathcal{C}} r_{ij}^\top \Sigma_{ij}^{-1} r_{ij}
\end{equation}

where $x$ is the parameter vector, $r_{ij}$ is the residual at $x$ associated with a constraint between variables $i$ and $j$ and $\Sigma_{ij}$ is the covariance matrix for the constraint. 
Alternatively, the quadratic form can be written as
\begin{equation} \label{eq:total-error-matrix}
    \footnotesize
    \text{arg}\min\limits_{x} \frac{1}{2}r^\top \Sigma^{-1}r.
\end{equation}
These problems may be expressed using an optimization framework~\cite{gtsam, g2o, Agarwal_Ceres_Solver_2022} and solved using algorithms such as Powell's dog leg or Levenberg-Marquardt.

\noindent\textbf{Levenberg-Marquardt for Nonlinear Least Squares:}
Given the nonlinear optimization problem in \autoref{eq:total-error}, the Levenberg-Marquardt algorithm (LM) can be used to iteratively find new parameters to drive the error towards a local minimum. LM modifies Gauss-Newton using a damping factor $\lambda$ to achieve better convergence. 
It exploits the first-order Taylor expansion
\begin{equation} \label{eq:taylor}
\footnotesize
f(x+\Delta x) \approx f(x) +  \frac{\partial f(x)}{\partial x} \Delta x.
\end{equation}

Each iteration then solves
\begin{equation} \label{eq:step-minimizer}
\footnotesize
\text{arg}\min\limits_{\Delta x} \frac{1}{2}||r + J\Delta x ||_{\Sigma}^{2} + \frac{\lambda}{2} ||D \Delta x||_2^2 
\end{equation}
at the current linearization point, where $J=\frac{\partial r}{\partial x}$, the Jacobian matrix, and $\frac{\lambda}{2} ||D \Delta x||_2^2$ is a regularization term to control the step size 
where $D^{\top}D$ is $\text{diag}(J^\top \Sigma^{-1} J)$ or $I$.
Taking the gradient of \autoref{eq:step-minimizer} with respect to $\Delta x$, setting it to 0, and rearranging the result gives the normal equations
\begin{equation} \label{eq:gradient}
\footnotesize
(J^\top \Sigma^{-1} J + \lambda D^{\top}D) \Delta x = -J^\top \Sigma^{-1} r.
\end{equation}

This can be more simply rewritten in terms of the regularized Hessian approximation $H_\lambda$ and gradient $b$ as
\begin{equation}  \label{eq:normal-equations}
\footnotesize
H_\lambda\Delta x = -b.
\end{equation}
Solving this linear system in each iteration yields $\Delta x$, which is a candidate parameter step to reduce the total error. Here, $H_\lambda$ is a sparse symmetric positive definite block matrix~\cite{Agarwal2010}.

\noindent\textbf{Visual-Inertial Optimization:}
Given a set of keyframes $\kappa$ and $l$ map points, the goal of visual bundle adjustment is to find new parameters that better satisfy camera reprojection error constraints. The cost may be formulated as \autoref{eq:visual-opt} ~\cite{ORBSLAM3} 

\begin{equation} \label{eq:visual-opt}
    \footnotesize
    \sum_{j=0}^{l-1}    \sum_{i\in \kappa ^ j} \rho_{\text{Huber}} (\|r_{ij}\|_{\Sigma_{ij}}^2)
\end{equation}

where $r_{ij}$ is the residual vector of the reprojection constraint between observed map point $j$ and keyframe $i$. 
In other words, $r_{ij}$ is the difference, in 2D coordinates, between the computed camera projection of a map point in a keyframe image using current parameters and its measured observation. 
To reduce sensitivity to feature mismatches between keyframes and potential outliers, a robust Huber kernel $\rho_{\text{Huber}}$ is used~\cite{ORBSLAM3}. Visual-inertial bundle adjustment introduces additional terms to the optimization for inertial parameters. For k+1 keyframes, this is represented as~\cite{ORBSLAM3}

\begin{equation} \label{eq:visual-inertial-opt}
    \footnotesize
    \sum_{i=1}^{k} \|r_{I_{i-1,i}}\|_{\Sigma_{I_{i-1,i}}}^2 +  \sum_{i=1}^{k} \|r_{b_{i-1,i}}\|_{\Sigma_{b_{i-1,i}}}^2
\end{equation}

where $r_{I_{i-1,i}}$ is the inertial residual between consecutive keyframes, representing the discrepancy between preintegrated IMU measurements and the change computed from keyframe parameters. These parameters include rotation, velocity, and position. Similarly, $r_{b_{i-1,i}}$ is the accelerometer and gyroscope bias residual between consecutive keyframes.

\noindent\textbf{GPU Programming:}
Modern GPUs are based on massively parallel Single-Instruction, Multiple-Thread (SIMT) architecture, where threads execute common instructions in groups. CUDA~\cite{cuda} is a programming model that enables writing specialized functions, called \emph{kernels}, which can be dispatched onto a GPU in order to process a batch of similar tasks in parallel. CUDA uses a hierarchical execution model where threads are grouped into 3D thread blocks. Thread blocks are further organized into a grid. When a kernel is launched, threads and blocks are assigned identifiers which can be queried by a task to decide inputs, outputs, and behaviour. Threads can then communicate with each other using shared memory and built-in functions. Additionally, multiple kernels can be launched in different streams, where each stream is a sequence of commands, to allow their executions to overlap on the GPU.

\section{Related Work} \label{sec:related-work}

As our work proposes a GPU-based approach for non-linear least squares optimization for problems such as bundle adjustment, we review literature on these topics.

Although there are several versatile libraries for least squares optimization in robotics and computer vision~\cite{gtsam, g2o, Agarwal_Ceres_Solver_2022, Rosen2017, sola_wolf_2022, Martiros-RSS-22}, they are primarily CPU-based and generally limit GPU acceleration to the linear solver stage~\cite{Agarwal_Ceres_Solver_2022}. Our framework differs by performing the entire optimization on the GPU, to enable greater parallelization for variable and constraint operations and avoid data transfers between intermediate steps to improve performance.

There are also a number of GPU-accelerated solvers based on domain-specific languages (DSLs)~\cite{OpenOF, Devito2017, Mara2021} or Python machine learning frameworks~\cite{Huang2021, Yi2021, pineda_theseus_2023, wang2023pypose}. However, these approaches may result in code duplication, lack the expressiveness needed to model complex types, or involve explicit language interoperation, depending on the application. Our framework avoids this by allowing problems to be described directly using user-defined data types, preventing code duplication, and enabling easier prototyping.

Several prior works also explore specialized implementations for bundle adjustment on the GPU, applying direct or iterative solving techniques~\cite{2021megba, Wu2011, HanschR.2016MMOB, cudabundleadjustment, Cao2021, daba, Han2025}. 
While performant, these works target a specific type of visual bundle adjustment which only uses camera poses, 3D points, and binary constraints. Meanwhile, SLAM and odometry systems often add several new constraints by incorporating inertial information~\cite{ORBSLAM3, Qin2018}, where some constraints have as many as six variables. 
They may also perform other optimization tasks such as motion-only bundle adjustment and pose graph optimization~\cite{ORBSLAM3, Qin2018}. Our framework resolves this limitation by supporting user-defined data types along with unary, binary, and n-ary GPU constraints.

While prior GPU-accelerated solvers utilize code generation and leverage existing machine learning frameworks, we adopt a design inspired by libraries such as g$^2$o and Ceres Solver, to support the level of functionality needed for visual-inertial SLAM~\cite{ORBSLAM3} and odometry systems~\cite{Qin2018}, discussed further in \autoref{sec:evaluation}.
The CUDA C++ interface enables the ability to leverage existing data structures within these systems for optimization. This approach also enables in-place optimization on platforms with full unified memory support, so that variables need not be explicitly transferred. Additionally, we support the ability to mix higher and lower floating-point precisions to reduce GPU memory usage.

\section{Design} \label{sec:design}

\begin{figure}[t]
  \vspace{1.5mm}
  \centering
  \includesvg[inkscapelatex=false, width=\linewidth]{figures/architecture}
  \caption{An example Graphite workflow. The user first defines properties for their variables and constraints, then constructs a graph by creating descriptors, and finally optimizes their variables using Levenberg-Marquardt.}
  \label{fig:overview}
  \vspace{-20pt}
  % \vspace{-1.5mm}
\end{figure}

\library{} implements nonlinear least squares graph optimization on the GPU, based on the Levenberg-Marquardt algorithm (\autoref{sec:background}). Its design is guided by the need to minimize the amount of time required for optimization while also using as little GPU memory as possible, which is crucial for resource-constrained systems. An optimization problem is represented by a graph constructed from a set of descriptors, which are collections of optimizable vertices (variables) and unary, binary, and n-ary constraints. Descriptors provide a batching mechanism for the GPU and allow vertices and constraints to be  represented with user data types that are compatible with CUDA device code (\autoref{sec:batching}). This avoids the need to first transform variables into a solver-specific format for optimization, a step which is often required by other libraries. As is standard with other frameworks, \library{} supports fast analytic differentiation (\autoref{sec:analytic}) and convenient automatic differentiation (\autoref{sec:autodiff}) for computing Jacobian matrices used by the optimization algorithm. To perform the optimization process quickly and in a memory-efficient manner, \library{} also supports mixed-precision solving (\autoref{sec:mixed-precision}).

We show a running example of the workflow in \autoref{fig:overview} for a 2D point optimization problem, where a user adjusts 2D points stored in GPU-accessible memory to lie on the boundary of a circle. They first define static properties for their 2D points and circle equation constraints. Next, they create the corresponding descriptors and construct a graph. Lastly, the user passes the graph to the Levenberg-Marquardt algorithm along with desired options, optimizing the points in-place. We discuss this process with concrete examples next.

\subsection{Descriptor Batching Model} \label{sec:batching}

A typical approach to offloading work onto a GPU is to organize similar tasks into batches, and process tasks within a batch concurrently across numerous threads in order to take advantage of the massively parallel architecture.
However, existing graph-based optimizers store different types of vertices and constraints indiscriminately together in the graph. This is not ideal for batching work on the GPU, because each thread processing an item (a vertex or constraint) may branch and execute a different code path, which is detrimental to performance~\cite{cuda}.
Therefore, to enable a GPU-friendly approach for graph-based optimization, \library{} allows users to construct an optimizable graph from collections of vertices and constraints of the same type. This enables GPU threads to process batches of identical items, which minimizes branching. We represent these collections using descriptors, which we describe next with our point optimization example. 

We demonstrate how the user models the point optimization problem from \autoref{fig:overview} in \library{}, where the user selects single precision for the variables and constraints, and 16-bit brain floating point for the linear system.

First, the user defines static properties for their vertices (points) and constraints (based on the circle equation), which are necessary for creating descriptors, as shown in \autoref{fig:overview}.
At compile time, these properties are used to specialize each CUDA kernel, so that each thread carries out the same type of computation, improving performance.

For a vertex descriptor, these properties include information such as the underlying data type (class), the parameter block size, a method to get its parameter block, and a method to update each vertex for a new step. \autoref{listing:point-descriptor-traits} shows how the user defines the properties for a 2D point descriptor. They specify the underlying vertex data type as a $2\times 1$ Eigen~\cite{Eigen} matrix (Line 2, Line 7), the length of the parameter block as 2 (Line 6), the method to get the parameter block (Lines 9 to 13), and the update method (Lines 15-18).

\begin{lstlisting}[language=C++, caption=Properties of a vertex descriptor for a 2D point., label=listing:point-descriptor-traits]
// Point definition
template <typename T> using Point = Eigen::Matrix<T, 2, 1>;

// Traits for Point
template <typename T, typename S> struct PointTraits {
  static constexpr size_t dimension = 2;
  using Vertex = Point<T>;

  template <typename P>
  d_fn static void parameters(const Vertex &vertex, P *parameters) {
    Eigen::Map<Eigen::Matrix<P, dimension, 1>> params_map(parameters);
    params_map = vertex.template cast<P>();
  }

  d_fn static void update(Vertex &vertex, const T *delta) {
    Eigen::Map<const Eigen::Matrix<T, dimension, 1>> d(delta);
    vertex += d;
  }
};

// Type alias
template <typename T, typename S>
using PointDescriptor = VertexDescriptor<T, S, PointTraits<T, S>>;
\end{lstlisting}

Next, the user defines static properties for their constraints, including which vertex descriptors are involved, the data types (classes) of the observations and any non-optimizable data, the loss function used, and the differentiation mode (discussed in \autoref{sec:analytic} and \autoref{sec:autodiff}). These properties also specify how to compute the residual for a constraint, and how to compute the Jacobians when using analytic differentiation (\autoref{sec:analytic}).
\autoref{listing:constraint-descriptor-traits} shows this for the point optimization problem. The user specifies a residual of size 1 (Line 3), specifies the vertex descriptor and data types (Lines 4-6), chooses the default loss function (Line 7), and chooses to use automatic differentiation (Line 8), so no Jacobian evaluation function is specified. Additionally, Lines 10-16 tell the descriptor how to compute the residual based on the equation of a circle.

\begin{lstlisting}[language=C++, caption=Properties for the circle equation constraint descriptor., label=listing:constraint-descriptor-traits]
// Traits for the circle constraint
template <typename T, typename S> struct CircleTraits {
  static constexpr size_t dimension = 1;
  using VertexDescriptors = std::tuple<PointDescriptor<T, S>>;
  using Observation = T;
  using Data = Empty; // unused - not passed to error function
  using Loss = DefaultLoss<T, dimension>;
  using Differentiation = DifferentiationMode::Auto;

  template <typename D>
  d_fn static void error(const D *point, const T &obs, D *error) {
    const auto x = point[0]; // Note: A const Point<T>& can also be passed in.
    const auto y = point[1]; // The framework automatically determines  
    const auto r = obs;      // how to call the user's function.
    error[0] = x * x + y * y - r * r;
  }
};

template <typename T, typename S>
using CircleDescriptor = FactorDescriptor<T, S, CircleTraits<T, S>>;
\end{lstlisting}

We specify properties of descriptors in this way because it lets users declare the behaviour of their types and automatically generate optimized GPU code, without relying on runtime lookups (as observed in some CPU-based optimizers~\cite{g2o}) to decide how each item in the graph is processed. Following \autoref{fig:overview}, after defining these properties, the user creates descriptors in order to construct an optimizable graph.

The user first creates a vertex descriptor for the collection of points, as shown in \autoref{listing:point-descriptor}. 
Before doing so, they declare a graph with the desired precisions in Lines 1-3.
Then in Lines 4-6, they create the vertex descriptor, reserve memory for the desired number of vertices, and add the descriptor to the graph. Then from Lines 8-10, they add a pointer for each vertex to the descriptor, and associate it with an identifier. In creating this descriptor, the user defines a batch of vertices which is processed efficiently across many GPU threads.

\begin{lstlisting}[language=C++, caption=Creating a vertex descriptor for 2D points., label=listing:point-descriptor]
using FP = float; // vertex, constraint precision
using SP = __nv_bfloat16; // linear system precision 
Graph<FP, SP> graph; // Declare graph
auto point_desc = PointDescriptor<FP, SP>(); // Create descriptor
point_desc.reserve(num_vertices); // Reserve memory for num_vertices
graph.add_descriptor(&point_desc); // Add to graph

for (size_t vertex_id = 0; vertex_id < num_vertices; ++vertex_id) {
    point_desc.add_vertex(vertex_id, &points[vertex_id]); // Add points
}
\end{lstlisting}

Next, they create a descriptor for the constraints in \autoref{listing:circle-factor}. As in the previous example, they initialize the descriptor, reserve memory for the desired number of constraints (one constraint for each vertex), and add the descriptor to the graph (Lines 1-3). They initialize the default loss function on Line 4, and then create the desired number of constraints (Lines 6-8). When creating each constraint, the user specifies the identifiers of the vertices involved, the observation (the radius of the circle), the precision matrix (\verb|nullptr| defaults to the identity matrix), optional constraint data (unused),
and the loss function parameters. Again, by constructing this descriptor, the user enables each constraint to be processed concurrently and efficiently on the GPU.

\begin{lstlisting}[language=C++, caption=Creating a descriptor for constraints., label=listing:circle-factor]
auto circle_desc = CircleDescriptor<FP, SP>(&point_desc); // Create descriptor
circle_desc.reserve(num_vertices); // Reserve memory for constraints
graph.add_descriptor(&circle_desc); // Add descriptor to graph
const auto loss = DefaultLoss<FP, 1>(); // loss function

for (size_t vertex_id = 0; vertex_id < num_vertices; ++vertex_id) {
    circle_desc.add_factor({vertex_id}, radius, nullptr, Empty(), loss);
}
\end{lstlisting}

After the user finishes constructing the graph, they pass it to an optimization algorithm, along with various options, including the desired linear solver (discussed in \autoref{sec:mixed-precision}), optimizing the points in-place (\autoref{fig:overview}). 
For the point optimization example, this is shown in \autoref{listing:circle-optimization}.
% The user optimizes the graph for the point optimization example in \autoref{listing:circle-optimization}. 
First, the user decides to fix the value of the last point they created (Line 1) and decides to disable the constraint which was created for the third point by changing its level to 1, as in g$^2$o~\cite{g2o} (Line 2). Next, the user configures an iterative linear solver, by selecting an identity matrix for the preconditioner (Line 3) and then instantiating the solver (Line 4) with their desired parameters. They create one stream (Line 5), since each constraint only has one vertex. Lastly, they configure the Levenberg-Marquardt algorithm on Lines 7-12, and run the optimization on Line 13.

\begin{lstlisting}[language=C++, caption=Example of optimizing noisy 2D points., label=listing:circle-optimization]
point_desc.set_fixed(num_vertices - 1, true); // Set the last vertex as fixed
circle_desc.set_active(2, 0x1); // Disable third constraint for point 2
IdentityPreconditioner<FP, SP> preconditioner;
PCGSolver<FP, SP> solver(50, 1e-6, 10.0, &preconditioner); 
StreamPool streams(1); // chose 1 stream for 1 vertex per constraint

optimizer::LevenbergMarquardtOptions<FP, SP> options;
options.solver = &solver;
options.initial_damping = 1e-6;
options.iterations = 10;
options.optimization_level = 0;
options.streams = &streams;
optimizer::levenberg_marquardt<FP, SP>(&graph, &options);
\end{lstlisting}

\subsection{Analytic Differentiation} \label{sec:analytic}

In graph-based optimization, constraints may involve one or more vertices, each of which generate a Jacobian matrix.
A typical way to compute Jacobians quickly is by evaluating an analytic (closed-form) expression. \library{} allows Jacobians to be evaluated directly using analytic expressions defined in a user-written function. These expressions are derived by hand or generated using a symbolic toolkit~\cite{Martiros-RSS-22, Cross2025}.

A typical approach on the CPU is to compute each matrix for a constraint sequentially~\cite{g2o, Agarwal_Ceres_Solver_2022}. However, to take advantage of the many GPU threads available, \library{} attempts to compute each Jacobian concurrently in separate streams (described in \autoref{sec:background}).
Additionally, \library{} allows these Jacobians to be evaluated dynamically~\cite{Wu2011}, at the point of computation (e.g. matrix-vector multiplication), to enable matrix-free solving methods which use less memory.

\subsection{Automatic Differentiation} \label{sec:autodiff}

Another method to compute Jacobians is through forward mode automatic differentiation. 
For example, Ceres Solver computes the gradient of a residual function using dual numbers in the form $a+b\varepsilon$, where a value is represented by real component $a$, and the partial derivative is represented by infinitesimal component $b$, where $\varepsilon^2 = 0$~\cite{Agarwal_Ceres_Solver_2022}.
To extend this for multiple inputs, Ceres implements a data structure called a \emph{Jet}, which replaces the infinitesimal component with a vector~\cite{Agarwal_Ceres_Solver_2022}. Ren et. al~\cite{2021megba} extend this concept to the GPU by proposing \emph{JetVectors}, which adopt a structure of arrays data layout to enable coalesced memory transactions for better performance. 
However, this design is intended for optimization across multiple devices and may require more GPU memory than available on a single device~\cite{2021megba}. 

To handle environments with less available memory, we avoid allocating storage for dual numbers entirely.
Instead of using Jets or JetVectors, when computing the residual of a constraint, each kernel replaces the input parameters with vectors of dual numbers, one for each vertex.
Each CUDA thread then calculates the column of a Jacobian matrix (\autoref{fig:autodiff}), by computing the residual using these vectors. The column to be computed is changed by setting one partial to 1, and the others to 0.
As shown in \autoref{sec:evaluation}, this uses the same amount of memory as analytic differentiation, allowing Jacobians to be computed efficiently and conveniently without first needing to derive an expression.

\begin{figure}[t]
  \vspace{0.5mm}
  \centering
  \includesvg[inkscapelatex=false, width=0.75\linewidth]{figures/autodiff}
  \caption{A CUDA thread calculates $\frac{\partial f}{\partial x_3}(\vec{x})$ by replacing the input with a vector of dual numbers and writes it to the corresponding Jacobian column.} 
  \label{fig:autodiff}
  \vspace{-20pt}
  \vspace{-0.5mm}
\end{figure}

\subsection{Mixed-Precision Solver} \label{sec:mixed-precision}

To enable mixed-precision solving, which allows the optimization to use less memory, \library{} supports the customization of floating point precisions for vertices and the linear system (\autoref{eq:normal-equations}). Essentially, these two precisions can be customized according to trade-offs between convergence, runtime performance, and memory usage. A typical approach would be to use a higher precision (double or single precision) for the vertices to maintain numerical stability, and an equal or lower precision (single precision or brain floating point) for the linear system to reduce the amount of memory needed to store it~\cite{Qin_Dayal_2023}.
In contrast, existing solvers support only one precision at a time~\cite{2021megba}, or only allow casting the linear system to single precision~\cite{Agarwal_Ceres_Solver_2022}. As shown in \autoref{sec:evaluation}, using lower precisions greatly reduces the overall optimization time and the amount of GPU memory required.

As described in \autoref{sec:background}, each iteration of Levenberg-Marquardt optimization computes a parameter step $\Delta x$. To compute this quickly on the GPU, we apply preconditioned conjugate gradients (PCG) on the Hessian, with implicit evaluation using the Jacobian matrices~\cite{Wu2011}, since this does not require storing the Hessian, and comprises of algebraic operations which are largely amenable to parallelization. However, directly using PCG with lower precision types can result in poor convergence. To mitigate this, we adopt techniques which are known to help with solving larger systems. These include clamping values along the Hessian diagonal~\cite{Agarwal_Ceres_Solver_2022}, rescaling the columns of each Jacobian block according to the Hessian diagonal~\cite{Agarwal_Ceres_Solver_2022, Qin_Dayal_2023}, and normalizing the PCG residual in each iteration~\cite{Guo2025}.

In applying PCG, to improve convergence, we also use a block Jacobi preconditioner~\cite{Agarwal2010}, because it is relatively simple to compute and does not require much memory due to its sparse block-diagonal matrix structure. We have observed that storing this preconditioner at too low of a precision can also result in convergence issues. Thus, when a 16-bit precision type such as bfloat16 is specified for the linear system, we compute and store this preconditioner at single or double precision, according to the precision of the variables.

Additionally, since the PCG solver does not require storing the Hessian matrix, we avoid explicitly constructing sparse matrices for the optimization problem. Instead, when performing an operation such as matrix-vector multiplication, the kernel uses the structure of the graph itself. This eliminates additional memory and runtime costs associated with converting and storing matrices in different formats, although each kernel must now be aware of the graph structure and whether a variable is fixed or a constraint is inactive.

\section{Evaluation} \label{sec:evaluation}
\begin{table*}[t]
\vspace{1.5mm}
\begin{center}
\begin{threeparttable}[b]
\caption{Bundle adjustment results for BAL (desktop), including  mean squared error, time (seconds), and GPU memory (MiB).}
\begin{tabular}{ |c|r|l|c|c|c|c|c|c|c| } 
 \hline
  \textbf{BAL} & \textbf{Problem} & \textbf{Evaluation} & \textbf{Graphite} & \textbf{Graphite} & \textbf{Graphite} & \textbf{Ceres} & \textbf{MegBA} &\textbf{MegBA} &\textbf{DeepLM} \\ 
 \textbf{Problem} & \textbf{Size\tnote{1}} & \textbf{Metric} & \textbf{FP64} & \textbf{FP32} & \textbf{FP32-BF16} & \textbf{CGNR} & \textbf{FP64} &\textbf{FP32} &\textbf{PCG} \\ 
\hline
& 16 Img & MSE & 0.43 & 0.43 & 0.43 & 0.43 & 0.43 & 0.43 & 0.43 \\
Dubrovnik-16 & 22106 Pts &  Time & 0.22 & 0.11 & 0.12 & 1.64 & 0.17 & 0.16 & 3.88  \\
& 83718 Obs & GPU Memory  & 298 & 282 & 278  & 336  & 406 & 352 & 814  \\
\hline
 & 21 Img & MSE & 1.67  & 1.67 & 1.68 & 1.67  & 1.67 & 1.67 & 1.67 \\
Trafalgar-21 & 11315 Pts & Time & 0.11 & 0.069 & 0.068  & 0.82  & 0.098 & 0.088 & 1.61   \\
& 36455 Obs & GPU Memory & 280 & 274 & 270  & 318 &  326 & 298 & 764  \\
\hline
& 49 Img & MSE & 0.85 & 0.85  &  0.85 & 0.84 & 0.84 & 0.84 & 0.84  \\
Ladybug-49 & 7776 Pts & Time & 0.063 & 0.041 & 0.046 & 0.66 & 0.086 & 0.074 & 1.14 \\
& 31843 Obs & GPU Memory & 278 & 274 & 270  & 316 & 326 & 300 & 740  \\
\hline
 & 1778 Img & MSE & 0.67 & 0.67 &  0.67 & 0.67 & 0.66 & 0.67 & 0.67  \\
 Venice-1778 & 993923 Pts & Time & 6.00 & 2.80 & 2.87  & 43.71 & 6.96 & 6.15 & 41.67  \\
& 5001946 Obs &  GPU Memory & 1944 & 1190 & 950  & 2160 & 7810 & 4528 & 5708  \\
\hline
 & 4585 Img & MSE & 1.13 & 1.13 & 1.14 & 1.14 & 1.12 & 1.13 & 1.13  \\
Final-4585 & 1324582 Pts & Time & 9.62 & 4.69 & 3.75 & 73.49 & 13.19 & 9.82 & 43.11  \\
& 9125125 Obs & GPU Memory & 3102 & 1822 & 1386  & 3546 & 13882 & 7960 & 7158 \\
\hline
 & 13682 Img & MSE & 1.50 & 1.50 & 1.50 & 1.51 & OOM & OOM & 1.51  \\
Final-13682 & 4456117 Pts & Time & 26.60 & 11.65 & 11.67 & 210.64 & OOM & OOM & 55.62  \\
& 28987644 Obs & GPU Memory & 9298 & 5212 & 3828  & 10640 & OOM  & OOM & 15044 \\
\hline
\end{tabular}
\label{tab:mixed-precision}
 \begin{tablenotes}
   \item [1] Shows the number of camera images (Img), points (Pts), and observations (Obs).
 \end{tablenotes}
\end{threeparttable}
\end{center}
\vspace{-20pt}
\vspace{-1.5mm}
\end{table*}

\begin{table}[t]
\vspace{1.5mm}
\caption{Bundle adjustment across differentiation modes, including mean squared error, time (seconds), and GPU memory (MiB).}
\begin{center}
\begin{tabular}{ |c|l|c|c|c| } 
 \hline
 \textbf{Problem} & \textbf{Metric} & \textbf{Analytic} & \textbf{Dynamic} & \textbf{Auto} \\ 
\hline
 & MSE & 1.50 & 1.50 & 1.50 \\
 Final-13682 & Time & 26.60 & 417 & 38.81 \\
 FP64 & GPU Memory & 9298 & 3988 & 9298 \\
\hline
 & MSE & 1.50 & 1.50 & 1.50 \\
 Final-13682 & Time & 11.65 & 25.22 & 11.46 \\
 FP32 & GPU Memory & 5212 & 2556 & 5212 \\
\hline
 & MSE & 1.50 & 1.50 & 1.50 \\
 Final-13682 & Time & 11.67 & 24.48 & 11.27 \\
FP32-BF16 & GPU Memory & 3828 & 2500 & 3828 \\
\hline
\end{tabular}
\end{center}
\label{tab:diff-modes}
\vspace{-20pt}
\vspace{-1.5mm}
\end{table}
% 50 iterations, 100 pcg iterations, 1e-12 tolerance, and inf rejection ratio

\begin{table*}[!ht]
\vspace{1.5mm}
\begin{center}
\begin{threeparttable}[b]
\caption{Full-inertial bundle adjustment on ORB-SLAM3 maps, including the total $\chi^{2}$ error, time (seconds) and GPU memory (MiB).}
\begin{tabular}{ |c|c|r|l|c|c|c|c|c| } 
 \hline
 \textbf{TUM-VI} & \textbf{Initial $\chi^2$} & \textbf{Graph} & \textbf{Evaluation} & \textbf{g$^2$o} & \textbf{g$^2$o} & \textbf{Graphite} & \textbf{Graphite} & \textbf{Graphite}\\ 
  \textbf{Sequence} & \textbf{Error}  & \textbf{Size\tnote{1}}  & \textbf{Metric\tnote{2}} & \textbf{LDLT}  & \textbf{PCG} & \textbf{FP64} & \textbf{FP32} & \textbf{FP32-BF16}\\ 
\hline
 & & 3131 KF & Final $\chi^2$ & $3.0918  \times 10^5$ & $4.4965 \times 10^5$ &  $4.3850 \times 10^5$ &  $4.6842 \times 10^5$ &  $5.9817 \times 10^5$\\
 outdoors1 & $1.0803\times 10^6$ & 58886 MP  & Time & 1424.40 & 224.08  &  4.96 & 1.71  &  1.18\\
  Desktop & &  348548 Co & GPU Mem. & --- & --- &  678 & 580 & 542\\
\hline
 & & 2032 KF & Final $\chi^2$ & $2.9105 \times 10^5$ & $3.5100\times 10^5$ &  $3.5012 \times 10^5$ &  $3.4090 \times 10^5$ &  $4.8966 \times 10^5$\\
 outdoors2 & $7.3020 \times 10^5$ & 57641 MP  & Time & 698.68 & 147.57  & 2.80 &  3.56 &  1.10\\
  Desktop & &  338364 Co & GPU Mem. & --- & --- &  656 &  564  &  530\\
\hline
 & & 1595 KF & Final $\chi^2$ & $2.5617 \times 10^5$ & $3.0916 \times 10^5$ &  $3.0354 \times 10^5$ &  $3.0684 \times 10^5$&  $3.3927 \times 10^5$ \\
 outdoors3 & $6.4905 \times 10^5$ & 36536 MP  & Time & 2957.40 & 99.69  &  1.69 &  1.32 &  1.13\\
  Desktop &  &  272307 Co & GPU Mem. & --- & --- &  636 &  562 & 534 \\
\hline
%jetson below
 & & 2483 KF & Final $\chi^2$ & $2.2494 \times 10^5$ & $2.9363 \times 10^5$ &  $ 2.9112 \times 10^ 5$ &  $ 2.9348 \times 10^5 $ & $ 3.6397 \times 10^5$\\
 outdoors1 & $7.1589 \times 10^5$ & 48621 MP  & Time & 2411.40 & 512.58  & 122.61 & 93.10 & 27.90\\
  Orin Nano & &  278661 Co & GPU Mem. & --- & --- &  5673, 306 & 5462, 95 & 5433, 66 \\
\hline
 & & 1972 KF & Final $\chi^2$ & $ 2.1432 \times 10^5$ & $3.3585 \times 10^5$ &  $ 3.3731 \times 10^5$ &  $ 3.5705 \times 10^5$& $3.5889 \times 10^5 $\\
 outdoors2 & $7.3551 \times 10^5$ & 33831 MP  & Time & 1141.40 &  345.46 & 53.63 & 32.58 & 29.73 \\
  Orin Nano & &  213125 Co & GPU Mem. & --- & --- & 5128, 328 & 5031, 231 & 5017, 217 \\
\hline
 & & 1589 KF & Final $\chi^2$ & $ 1.6574 \times 10^5$ & $ 1.9508 \times 10^5$ &  $ 1.9439 \times 10^5$ &  $ 2.0736 \times 10^5 $ & $ 2.1015 \times 10^5$\\
 outdoors3 & $ 5.3520 \times 10^5$ & 24524 MP  & Time & 2166.1 & 355.34  &  93.71 & 33.53  & 29.64 \\
  Orin Nano & &  174070 Co & GPU Mem. & --- & --- & 4674, 282 & 4473, 81 & 4449, 57 \\
\hline
\end{tabular}
\label{tab:fiba}
 \begin{tablenotes}
   \item [1] Shows the number of keyframes (KF), map points (MP), and constraints (Co). Each keyframe generates four variables (pose, velocity, IMU biases).
   \item [2] For the Orin Nano, we report the peak GPU memory usage and peak usage minus the baseline usage before Graphite runs.
 \end{tablenotes}
\end{threeparttable}
\end{center}
\vspace{-20pt}
\vspace{-1.5mm}
\end{table*}

We evaluate \library{} in terms of the error reduction, runtime performance, and GPU memory usage. First, we compare the solver in different precision modes, using small to large problems from Bundle Adjustment in the Large (BAL)~\cite{Agarwal2010}, which serves as a common benchmark across optimization libraries.
Next, we compare analytic, dynamic, and automatic differentiation modes. Lastly, we run full-inertial bundle adjustment with \library{} inside ORB-SLAM3, using maps generated from outdoors sequences in TUM-VI~\cite{schubert2018vidataset}, a well-known dataset for evaluating visual-inertial SLAM systems. 
% We run these experiments using a desktop machine equipped with a discrete graphics card, and an NVIDIA Jetson Orin Nano Developer Kit, an embedded robotics platform, to assess the performance of \library{} in high-performance and resource-constrained environments (specifications in \autoref{tab:specs}). 
To assess \library{} in high-performance and resource-constrained environments, we run these experiments on a desktop machine with a 12-core Intel Core i7-12700K CPU at 3.8-4.9 GHz, 10496-core NVIDIA RTX 3090 GPU at 1.9 GHz, and 64 GB RAM, and also on an NVIDIA Jetson Orin Nano with a 6-core ARM Cortex-A78AE CPU at 1.7 GHz, 1024-core NVIDIA Ampere GPU at 1.0 GHz, and 8 GB RAM. We use unified memory to store optimizable variables.

\subsection{Mixed-Precision Modes}

We evaluate \library{} on BAL datasets using double precision (FP64), single precision (FP32), and single precision mixed with bfloat16 (FP32-BF16) on the desktop machine. We use bfloat16 rather than 16-bit half precision, because it can approximately represent the same range as single precision.
For comparison, we select two general GPU-accelerated optimization libraries with BAL support and similar conjugate gradient solvers, Ceres Solver~\cite{Agarwal_Ceres_Solver_2022} and DeepLM~\cite{Huang2021}, as well as MegBA~\cite{2021megba}, a fully-accelerated, high-performance bundle adjuster with single and double precision modes.
Where available, we use analytic Jacobians, which have the least runtime overhead. We generate Jacobian expressions for \library{} using wrenfold~\cite{Cross2025}.
To allow each optimizer sufficient opportunity to converge, we use a maximum of 50 iterations.
For PCG in MegBA, the maximum number of iterations is set to 100, while 10 is used for the other configurations, since they solve the full Hessian system rather than a reduced one, which involves more work per iteration.
As shown in \autoref{tab:mixed-precision} for single run experiments, \library{} uses substantially less memory when using BF16 to store the Jacobian matrices, at the cost of slightly worse convergence. For runtime, FP64 takes the longest, while FP32 and FP32-BF16 take a comparable amount of time. Compared to similar libraries, \library{} takes less time since it accelerates constraint-specific calculations as well.
Although MegBA achieves slightly lower MSE in a few instances, \library{} uses $\frac{1}{4}$ the memory for large problems, since it avoids storing Hessian matrices, and receives a greater speedup when using single precision. These results highlight the runtime and memory advantages of performing the entire optimization on the GPU and directly utilizing the graph structure to avoid constructing intermediate matrices.

\subsection{Differentiation Modes}

We report \library{}'s performance using analytic, dynamic, and automatic differentiation modes in \autoref{tab:diff-modes} for Final-13682, the largest dataset in BAL. Automatic differentiation allocates the same amount of memory as analytic differentiation, 
taking approximately the same time at lower precisions,
and significantly more at FP64. Meanwhile, the dynamic mode requires the least amount of GPU memory since Jacobians are not stored, although precision matrices are still used. While the dynamic mode exhibits a larger slowdown for FP64, this can be mitigated by switching to lower precisions. Our results show that automatic differentiation can be achieved with relatively low memory and runtime overhead for lower precisions, enabling rapid prototyping of optimizable constraints, while dynamically computed Jacobians enable more memory efficient solving at the cost of higher runtimes.

\subsection{Case Study: Visual-Inertial Bundle Adjustment}

We apply \library{} to full-inertial bundle adjustment in ORB-SLAM3, which introduces several additional variables and constraints to visual bundle adjustment (\autoref{sec:background}).

\noindent\textbf{Implementation:} To use \library{} for visual-inertial bundle adjustment, we reimplement vertices and constraints involved in the optimization, external to the library.
The original data types in ORB-SLAM3 cannot be used as-is, since some contain extraneous data, involve dynamic memory allocations which must be manually transferred from the CPU to the GPU, or implement methods which are not suitable for GPU execution (e.g. a method which accesses a resource shared between different CPU threads).
Therefore, we implement lightweight template data types as a replacement, which contain only the subset of data and functionality needed for visual-inertial optimization. These include data types for the combined IMU camera pose vertex, the cameras (pinhole and fisheye), and the preintegrated IMU measurements.
The remaining vertex data types for the map points, velocities, gyro biases, and accelerometer biases are $3\times 1$ vectors, so we directly represent these using Eigen~\cite{Eigen}. To use these data types with \library{}, we define descriptors, as in \autoref{listing:point-descriptor-traits}.

To model the optimization constraints, we create eight descriptors for monocular and stereo reprojection errors, the inertial residual (with and without Huber loss), gyro and accelerometer bias residuals, and bias priors. 
Following the original implementation, we use the same expressions for computing the residuals and Jacobians of each constraint along with Sophus~\cite{sophus} for Lie groups.
When constructing the graph, to avoid overhead from reallocating memory, we predetermine the size of the optimization and reserve the required memory for descriptors and optimizable variables.

\noindent\textbf{Experiment:} Next, we evaluate the performance of visual-inertial bundle adjustment inside ORB-SLAM3 using Graphite. Among compatible datasets, we choose outdoors sequences from the TUM-VI dataset, since they generate large maps consisting of thousands of keyframes. 
Within ORB-SLAM3, we perform full (global) visual-inertial bundle adjustment on each of the generated maps.
For the baseline, we compare \library{} to the existing LDLT-based g$^2$o implementation, using the original LM algorithm without modified termination criteria~\cite{ORBSLAM3} for more consistent behaviour. We also include a PCG baseline, since our implementation uses PCG. Note that the challenges discussed in \autoref{sec:related-work} restrict which solvers can be used. Limiting factors include language compatibility and sufficient support for user data types involved in complex constraints (e.g. with up to six variables).
For all solvers, the parameter recovery step is skipped so that all optimizations run on the same map. The optimization is set to run for up to 100 iterations. Block ordering is enabled for LDLT as we found this improves the runtime performance. For g$^2$o PCG, we use a maximum of 50 iterations, with $1\times10^{-6}$ absolute tolerance, while for our solver, we use a maximum of 100 iterations, with $1\times10^{-12}$ absolute tolerance, and a rejection ratio of 10. The reason for different termination criteria is that g$^2$o computes and solves a reduced system, while \library{} uses the full Hessian and gradient. Both solvers apply block Jacobi preconditioning.

\noindent\textbf{Results:} The results for TUM-VI are given in \autoref{tab:fiba}. 
The LDLT baseline converges to the lowest $\chi^2$ error for all maps due to computing the exact parameter step at each iteration, at the cost of extremely high runtimes. Compared to the PCG baseline, we achieve a relative speedup of up to $59 \times$ on the desktop and up to $6 \times$ on the Orin Nano, despite using the full Hessian. 
This difference in speedup is in part due to the Orin Nano having far fewer CUDA cores and operating at lower clock speeds than its desktop counterpart. 
Moreover, while lower precision modes take less time on the Jetson, FP64 generally achieves lower $\chi^2$ values and runs for more iterations. Note that the Jetson reports higher memory usages due to the different architecture, so we also include the peak usage minus the baseline usage before Graphite first runs.
Overall, \library{} achieves a significant reduction in time, enabling faster large-scale visual-inertial optimization.

\section{Conclusion}

We have implemented a general GPU-accelerated nonlinear optimization framework based on Levenberg-Marquardt for sparse estimation problems in CUDA C++, which enables the sharing of data types and functions between real-time applications and optimization tasks. \library{} achieves a relatively lower memory footprint by supporting in-place optimization, inferring matrix structures from the optimizable graph, supporting mixed-precision solving, and enabling matrix-free methods. Our results show that our framework enables large-scale optimization over complex constraints for desktop and resource-constrained devices. In the future, we would like to support more algorithms and solving techniques.

% uncomment if needed when finalizing paper - it causes references to be split across multiple pages

%%%%%%%%%%%%%%%%%%%%%%%%%%%%%%%%%%%%%%%%%%%%%%%%%%%%%%%%%%%%%%%%%%%%%%%%%%%%%%%%

%%%%%%%%%%%%%%%%%%%%%%%%%%%%%%%%%%%%%%%%%%%%%%%%%%%%%%%%%%%%%%%%%%%%%%%%%%%%%%%%



% \section*{ACKNOWLEDGMENT}

% The preferred spelling of the word ÒacknowledgmentÓ in America is without an ÒeÓ after the ÒgÓ. Avoid the stilted expression, ÒOne of us (R. B. G.) thanks . . .Ó  Instead, try ÒR. B. G. thanksÓ. Put sponsor acknowledgments in the unnumbered footnote on the first page.

%%%%%%%%%%%%%%%%%%%%%%%%%%%%%%%%%%%%%%%%%%%%%%%%%%%%%%%%%%%%%%%%%%%%%%%%%%%%%%%%

\bibliographystyle{IEEEtran}
\bibliography{IEEEabrv,references}

\end{document}